\documentclass[10pt,twocolumn,letterpaper]{article}

\usepackage[pagenumbers]{cvpr} 

\usepackage{algorithm}
\usepackage{algorithmic}

\definecolor{cvprblue}{rgb}{0.21,0.49,0.74}
\usepackage[pagebackref,breaklinks,colorlinks,allcolors=cvprblue]{hyperref}

\def\paperID{16377} 
\def\confName{CVPR}
\def\confYear{2025}

\begin{document}
\title{Structure-Preserving Zero-Shot Image Editing \\ via Stage-Wise Latent Injection in Diffusion Models
}

\author{Dasol Jeong$^1$ \quad Donggoo Kang$^1$ \quad Jiwon Park$^2$ \quad Hyebean Lee$^1$ \quad Joonki Paik$^{1,2,\star}$ \\
$^1$Department of Image and $^2$Department of Artificial Intelligence \\ Chung-Ang University\\
}


\maketitle
\let\thefootnote\relax\footnote{This work has been submitted to the IEEE for possible publication. Copyright may be transferred without notice, after which this version may no longer be accessible.
}

\begin{abstract}
We propose a diffusion-based framework for zero-shot image editing that unifies text-guided and reference-guided approaches without requiring fine-tuning. Our method leverages diffusion inversion and timestep-specific null-text embeddings to preserve the structural integrity of the source image. By introducing a stage-wise latent injection strategy—shape injection in early steps and attribute injection in later steps—we enable precise, fine-grained modifications while maintaining global consistency. Cross-attention with reference latents facilitates semantic alignment between the source and reference. Extensive experiments across expression transfer, texture transformation, and style infusion demonstrate state-of-the-art performance, confirming the method's scalability and adaptability to diverse image editing scenarios.
\end{abstract}    
\section{Introduction}
\label{sec:intro}

Image editing is a fundamental task in computer vision, with applications in digital art, content creation, and scientific visualization~\cite{rombach2022high, meng2021sdedit, hertz2022prompt}. Traditional editing methods often require manual intervention and domain-specific expertise, limiting accessibility and scalability. Recent advancements in diffusion models have enabled high-quality, automated image editing, but existing approaches still face limitations in structural fidelity and flexible attribute control.

\begin{figure}[h]
    \centering
    \includegraphics[width=\linewidth]{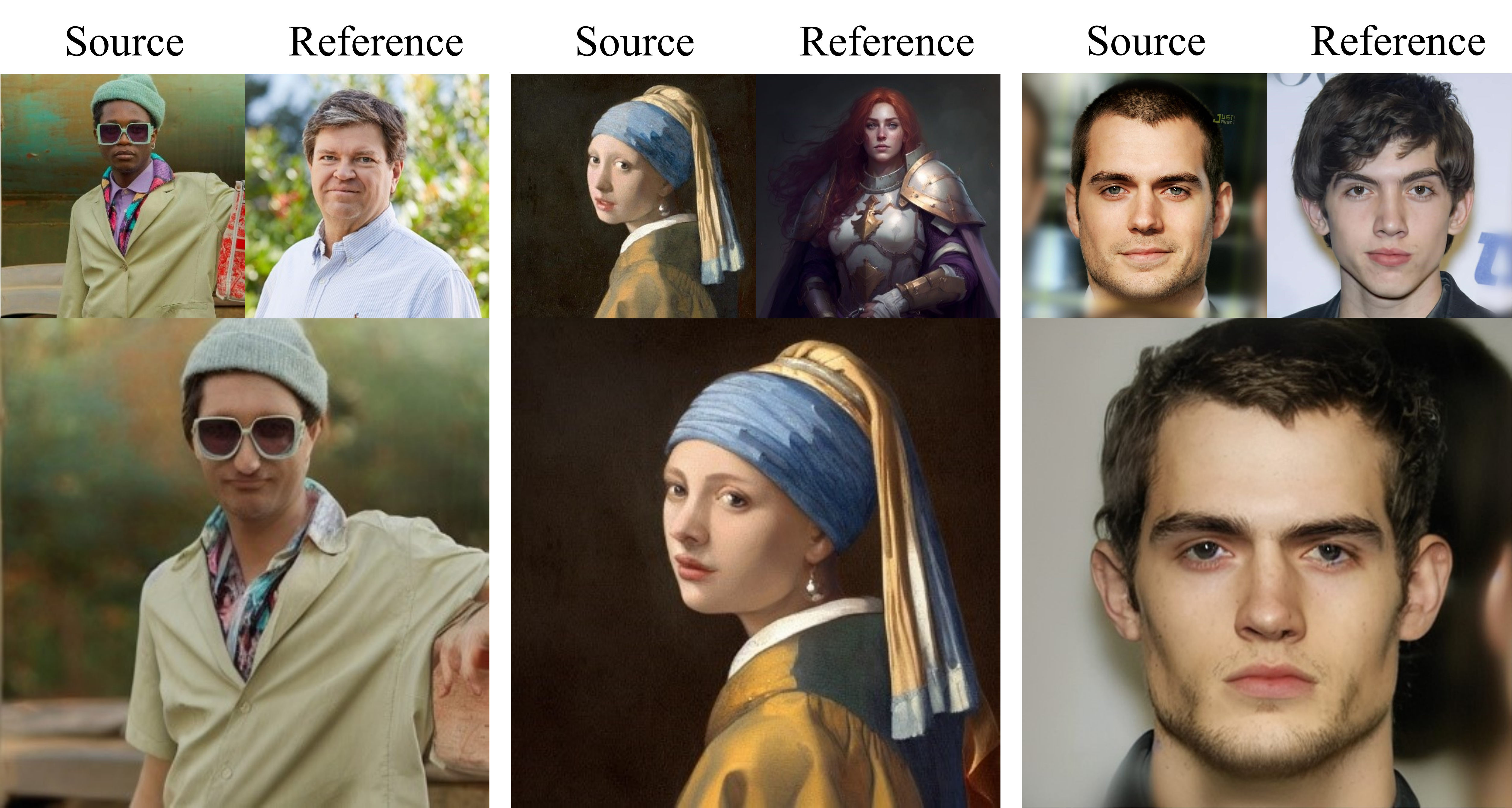}
    \caption{Results of reference-guided image editing using the proposed method.}
    \label{fig:reference_edit_person}
\end{figure}

Text-guided image editing, powered by models such as CLIP~\cite{radford2021learning}, allows users to modify images based on natural language descriptions. While intuitive, these methods~\cite{avrahami2022blended, nichol2021glide, kim2022diffusionclip} often lack fine-grained control and fail to preserve the structural integrity of the source image, particularly in localized modifications. Conversely, reference-guided editing~\cite{ruiz2023dreambooth, dong2022dreamartist, gal2022image, hu2021lora} transfers attributes from an example image but frequently distorts the original structure and requires fine-tuning for diverse tasks. These challenges highlight the need for a unified framework that can effectively combine text-based and reference-based modifications while ensuring structural consistency.

A widely used technique in real image editing is DDIM inversion~\cite{song2020denoising}, which reconstructs the latent representation of an image by reversing the denoising process. This allows generative models to manipulate real images in a controlled manner~\cite{mokady2023null, miyake2023negative, han2024proxedit, zhang2024real, cao2023masactrl, elarabawy2022direct}.

A promising direction is Null-Text Inversion (NTI)~\cite{mokady2023null}, which enables text-guided editing while preserving the structure of real images. However, it is inherently limited to textual modifications and does not support reference-based attribute transfer. Similarly, RIVAL~\cite{zhang2024real} improves inversion-based structural alignment but struggles with localized fine-grained attribute control, leading to unwanted distortions when transferring styles or textures from reference images. These challenges necessitate an approach that seamlessly integrates text and reference guidance while preserving structural integrity.

To address these limitations, we propose a zero-shot image editing framework that unifies text-guided and reference-guided methods using a stage-wise injection strategy within a diffusion-based architecture. Our approach introduces self-attention on the reference image in the early timesteps to preserve spatial structure, ensuring that the edited output maintains the form and layout of the source image. As the denoising process progresses, attribute injection via self-attention on the reference latents enables fine-grained semantic and stylistic transfer, maintaining coherence with the source image while ensuring that the transferred attributes are contextually aligned rather than arbitrarily imposed.

Fig.~\ref{fig:reference_edit_person} illustrates the effectiveness of our approach, demonstrating how our method seamlessly integrates desired modifications without the need for masks, fine-tuning, or additional constraints. Unlike prior methods that require explicit region annotations or heavy architectural modifications, our framework accurately transforms only the intended regions while preserving the global structure of the image. The results highlight the robustness of our stage-wise shape and attribute injection, ensuring that fine-grained transformations are both structurally coherent and semantically meaningful.

Additionally, our approach optimizes null-text embeddings to act as a structural reference, preventing unwanted deformations during editing. Unlike existing methods that treat null-text embeddings as simple placeholders, we dynamically adjust them based on diffusion timesteps, ensuring that structural consistency is maintained while enabling controlled semantic modifications. By leveraging this combination of structural preservation and fine-grained control, our framework achieves high-quality zero-shot image editing without requiring task-specific fine-tuning.

Extensive experiments on AFHQ\cite{choi2020starganv2} and Oxford-IIIT Pets\cite{parkhi12a} demonstrate state-of-the-art performance in perceptual quality, semantic consistency, and structural preservation. Our method outperforms existing approaches in both zero-shot reference- and text-guided editing tasks, establishing a new benchmark for diffusion-based image editing.
\section{Related Work}
\label{sec:relatedwork}

\subsection{Diffusion Models for Image Editing}
Diffusion models have demonstrated remarkable capabilities in generating high-quality and semantically coherent images~\cite{ho2020denoising, song2020score, rombach2022high, podell2023sdxl}. 
By operating in a compressed latent space, these models have significantly improved the efficiency and scalability of image synthesis and editing.

Text-guided image editing~\cite{meng2021sdedit, saharia2022image} has achieved remarkable progress, leveraging models like CLIP for intuitive natural language-based manipulation. Despite these advancements, such methods often lack fine-grained control and struggle to preserve structural integrity during complex edits, limiting their applicability in tasks requiring precise transformations.

Reference-guided image editing~\cite{gal2022image, ruiz2023dreambooth, dong2022dreamartist} provides strong attribute transfer capabilities but frequently distorts structural integrity, especially when transferring complex features such as textures or expressions. Many of these methods require extensive fine-tuning, making them less practical for zero-shot scenarios.

To enhance control, diffusion-based generative models have increasingly incorporated mask-guided techniques~\cite{avrahami2022blended, nichol2021glide, hertz2022prompt} for tasks such as inpainting, attribute manipulation, and structure-preserving editing. However, these approaches often rely on user-specified masks for precise control, introducing additional complexity and reducing their practicality in zero-shot settings.

\subsection{DDIM Inversion for Real Image Editing}

A key challenge in real image editing is reconstructing an image in the diffusion latent space while preserving its original structure. DDIM Inversion~\cite{song2020denoising, dhariwal2021diffusion} enables this by reversing the forward diffusion process, allowing fine-grained edits while maintaining the global structure.

Null-Text Inversion (NTI)~\cite{mokady2023null} builds upon DDIM inversion to preserve structure during text-guided editing. By optimizing null-text embeddings, NTI allows controlled semantic modifications without requiring fine-tuning. However, it is inherently limited to text-based prompts and cannot integrate external reference attributes.

RIVAL~\cite{zhang2024real} extends DDIM inversion to better align structural representations across real images. While effective at handling domain shifts, RIVAL still struggles with fine-grained attribute transfer, particularly in localized modifications such as expression changes or texture blending.

Although these methods improve real image editing, they remain either constrained to text guidance (NTI) or prone to distortions in reference-guided tasks (RIVAL). This limitation motivates our stage-wise injection framework, which preserves structural consistency while enabling precise attribute transfer.

\subsection{Image-to-Image Translation in Diffusion Models}

Recent diffusion-based image-to-image translation methods~\cite{kwon2022diffusion, wu2023latent, xia2024diffi2i} have introduced techniques for controlling content and style separately. While these models show strong performance in reference-guided editing, they often require paired training data or additional fine-tuning to generalize across diverse image domains.

InjectFusion\cite{jeong2024training} proposes a fusion mechanism for content transfer, but its results tend to be over-smoothed, leading to a loss of high-frequency details.
DiffuseIT\cite{kwon2022diffusion} disentangles style and content for improved reference-based editing but fails to consistently preserve geometric structures.

In contrast, our approach does not require fine-tuning and introduces a two-stage latent injection mechanism that allows localized semantic modifications without compromising global structure.

\section{Method}

Our proposed framework leverages diffusion-based generative models to achieve zero-shot image editing with both structural fidelity and semantic adaptability. The method comprises three key components: (1) source image inversion with optimizing a null-text embedding and reference image inversion, (2) shape and attribute injection during diffusion sampling, and (3) classifier-free guidance for balancing structure and style. Below, we describe each component in detail.

\begin{figure}[t]
\vskip 0.2in
\begin{center}
\centerline{\includegraphics[width=0.9\columnwidth]{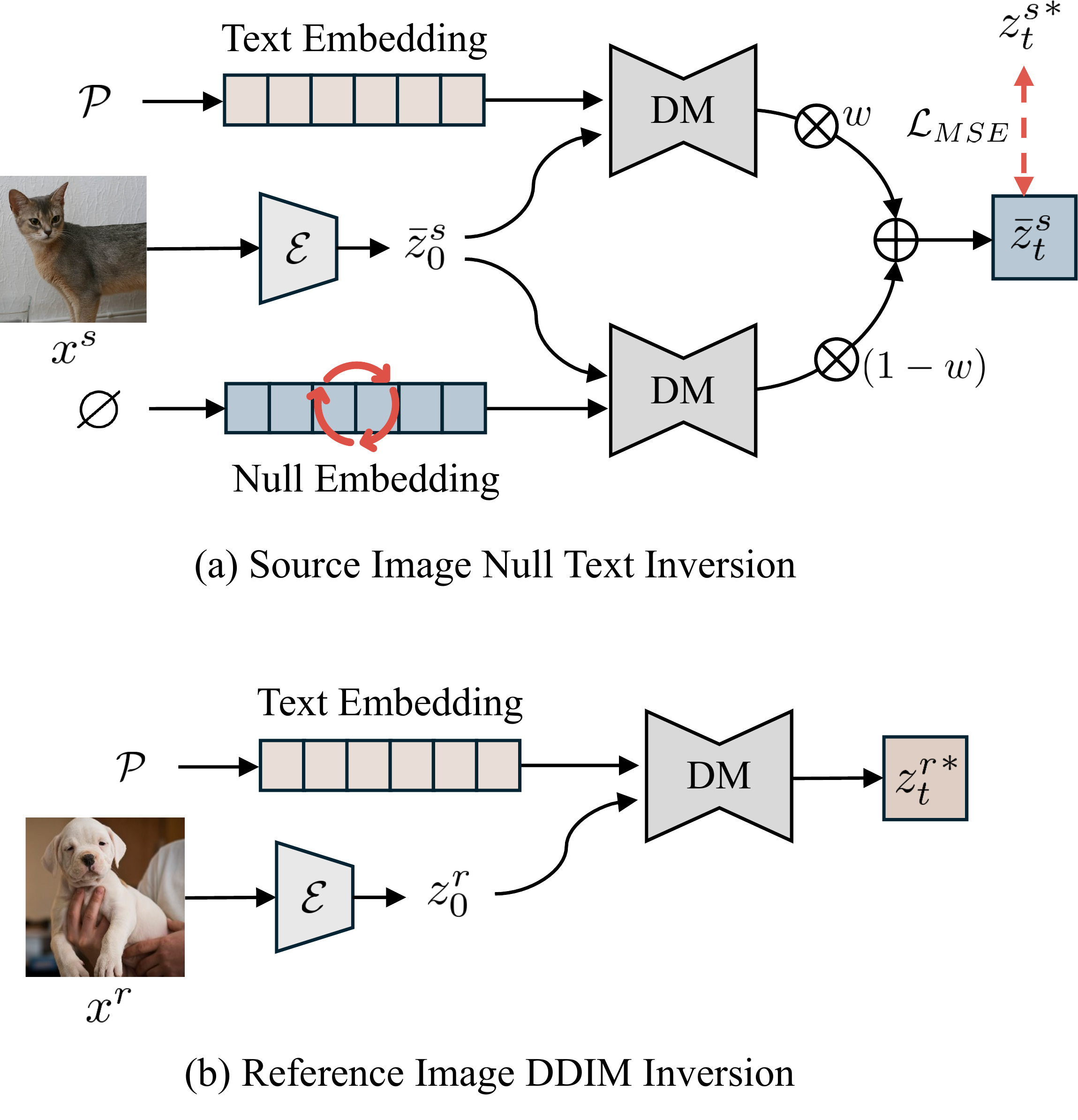}}
\caption{Inversion process for source and reference images.
(a) Source Image Null-Text Inversion: The source image undergoes Null-Text Inversion (NTI) to optimize null-text embeddings, ensuring that its structural integrity is preserved throughout the editing process. This allows for precise text- or reference-guided modifications without fine-tuning.
(b) Reference Image DDIM Inversion: The reference image is inverted using DDIM inversion, generating a latent representation that captures fine-grained attributes such as texture, style, and expression. These reference latents are later integrated into the denoising process to guide attribute injection.}
\label{inversion}
\end{center}
\vskip -0.2in
\end{figure}

\subsection{DDIM Inversion}
To combine the source and reference images in the latent space, we perform DDIM inversion~\cite{song2020denoising, song2020score} to generate inverted latent representation sets $\{z_t^{s*}\}_{t=1}^T$ and $\{z_t^{r*}\}_{t=1}^T$ respectively. 
The inverse process of DDIM sampling~\cite{dhariwal2021diffusion} is performed using the stable diffusion model~\cite{rombach2022high}, where the guidance scale $w$ is set to 1 in the classifier-free guidance~\cite{ho2022classifier}.

To preserve the structural properties of the source image and improve the reconstruction quality at each timestep, we adapt the Null-text Inversion~\cite{mokady2023null}. 
Fig.~\ref{inversion}, (a) optimizes null embeddings $\varnothing_t$ to replace the null prompt embedding $\varnothing = \psi("")$ in classifier-free guidance, preserving the timestep-specific properties of the source image without fine-tuning. 
The optimization of the null text embedding using the previously generated source latent representation set $\{z_t^{s*}\}_{t=1}^T$ and the conditional prompt embedding $\mathcal{C} = \psi(\mathcal{P})$ is expressed as:

\begin{figure*}[h]
\vskip 0.2in
\begin{center}
\centerline{\includegraphics[width=0.95\linewidth]{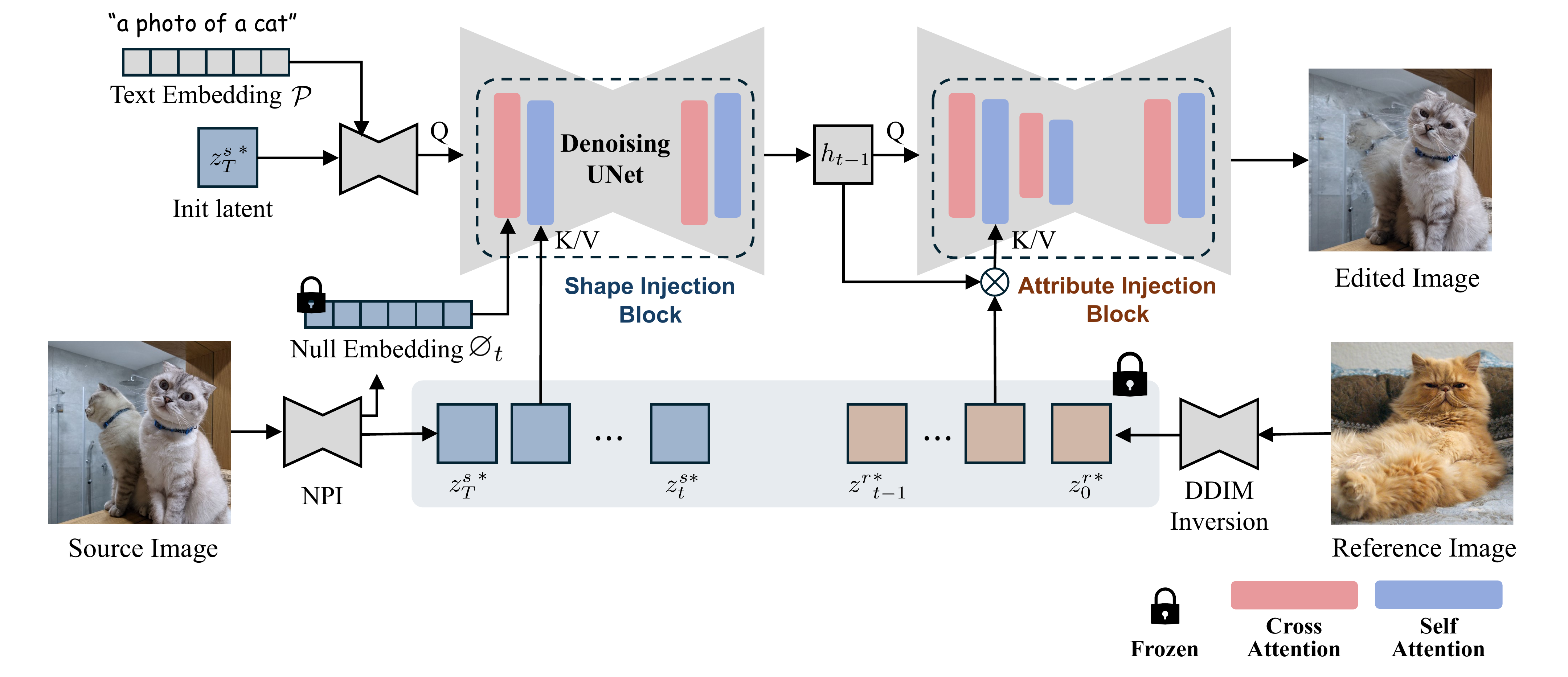}}
\caption{\textbf{An overview of the proposed zero-shot image editing framework.} The process begins with DDIM inversion of the source image to extract its latent representation $z_t^{s*}$ and optimize null embeddings $\varnothing$ for structural preservation. Text embeddings $\mathcal{P}$ and reference image latents $z_t^{r*}$ guide the denoising U-Net during the editing process. The framework incorporates shape injection during early timesteps to maintain structural fidelity and attribute injection in later timesteps to transfer fine-grained semantic and stylistic attributes from the reference image. Self-attention and cross-attention mechanisms within the U-Net enable precise control over structural and semantic transformations, ensuring high-quality editing results.}
\label{overview}
\end{center}
\vskip -0.2in
\end{figure*}

\begin{equation}\label{eq:source_inversion}
    \min_{\varnothing_t} ||z_{t-1}^{s*} - \epsilon_{t-1}(\bar{z}^s_t, \varnothing_t, \mathcal{C})||^2_2 
\end{equation}

where $\epsilon_{t-1}$ denotes the DDIM sampling step, and the guidance scale $w$ is set to 7.5.
The optimized null text embeddings $\{\varnothing_t\}_{t=1}^T$ serve as anchors for structural consistency during the editing process.




\subsection{Shape Injection}
RIVAL~\cite{zhang2024real} transforms the source image by injecting the current latent generated as a conditional prompt into the source latent within the attention layer of the U-Net architecture of the diffusion model.
Inspired by this, we propose an integrated injection mechanism that injects a reference image into the source image.
This mechanism consists of Shape and Attribute Injection blocks, as shown in the figure. ~\ref{overview}.
Self-attention and cross-attention within each block follow the attention expression~\cite{vaswani2017attention}, $\text{softmax} (Q K^\top \slash \sqrt{d}) V$.


The shape injection block ensures that the spatial structure of the source image is preserved during the early timestep (\(t \leq t_{\text{early}}\)). During the early timestep, the query, key, and value of the self-attention are defined as follows:

\begin{equation}\label{eq:attention}
    Q = W_Q(h_t), K = W_K(z^{s*}_t), V = W_V(z^{s*}_t),
\end{equation}

where each $W$ represents a learned linear projection, and $h_t$ is the query latent representation.

The optimized null text embedding $\varnothing_t$ and the conditional prompt embedding $\mathcal{C}$ perform Cross-attention on the current query latent representation $h_t$, respectively.
This is used in classifier-free guidance to balance structure preservation and semantic adaptability, and the equation is as follows:

\begin{equation}\label{eq:cfg_shape}
    \tilde{\epsilon}_t = w \cdot \epsilon_\theta(z^{s*}_t, t, \mathcal{C}) + (1 - w) \cdot \epsilon_\theta(z^{s*}_t, t, \varnothing_t),
\end{equation}

where $\varnothing_t$ acts as an unconditional prompt instead of a null prompt, minimizing semantic damage at each timestep and improving the reconstruction quality.

\subsection{Attribute Injection}

As the denoising process progresses (\(t > t_{\text{early}}\)), the attribute injection stage integrates stylistic and semantic attributes from the reference image into the source latent. Cross-attention mechanisms guide this process, using the reference latent, $z^{r*}_t$, as the key and value to inject fine-grained details such as expressions and textures:

\begin{equation}\label{eq:attention}
    Q = W_Q(h_t), K = W_K(z^{r*}_t \oplus h_t), V = W_V(z^{r*}_t \oplus h_t),
\end{equation}

where $\oplus$ represents connectivity, which can accurately integrate reference properties while preserving the spatial structure of the source image.

During this timestep, the unconditional prompt is given a null prompt embedding$\varnothing = \psi("")$, and the conditional prompt embedding performs the same cross-attention to perform classifier-free guidance as follows:

\begin{equation}\label{eq:cfg_attr}
    \tilde{\epsilon}_t = w \cdot \epsilon_\theta(z^{r*}_t, t, \mathcal{C}) + (1 - w) \cdot \epsilon_\theta(z^{r*}_t, t, \varnothing),
\end{equation}

By combining these components, our method achieves a balanced and flexible editing pipeline, preserving the source image's structural shape while seamlessly integrating fine-grained attributes from the reference image. Algorithm~\ref{alg:injection} outlines the complete editing pipeline, detailing how the source and reference latents are combined during the diffusion process.

\begin{algorithm}[tb]
   \caption{Reference-guided Image Editing}
   \label{alg:injection}
\begin{algorithmic}
   \STATE {\bfseries Input:} Source image $x^s$, reference image $x^r$, source prompt embedding  $\mathcal{C_\text{src}}$, target prompt embedding $\mathcal{C_\text{edit}}$, pretrained model $\epsilon_\theta$
   \STATE {\bfseries Output:} Edited image $x_{\text{edit}}$

   \STATE $\{z_t^{s*}\}^T_{t=1}, \{\varnothing_t\}^T_{t=1} \leftarrow NTI(x^s, \mathcal{C_\text{src}});$
   \STATE $\{z_t^{r*}\}^T_{t=1} \leftarrow DDIMinv(x^r);$

   \STATE Initialize $z_T \leftarrow z_T^*;$
   \FOR{$t=T, \dots ,1$}
   \IF{$t \geq t_{early}$} {
   \STATE $z_{t-1} \leftarrow \epsilon_\theta(z_t, z_t^{s*}, \varnothing_t, \mathcal{C_\text{edit}})$

   \STATE $\tilde{\epsilon_t} = (1-\alpha) \cdot \epsilon_\theta(z_t^{s*}, t, \varnothing_t) + \alpha \cdot \epsilon_\theta(z_t, t, \mathcal{C_\text{edit}})$
   }

   \ELSE {
   \STATE $z_{t-1} \leftarrow \epsilon_\theta(z_t, z_t^{r*}, \mathcal{C_\text{edit}})$
   \STATE $\tilde{\epsilon_t} = (1-\alpha) \cdot \epsilon_\theta(z_t^{r*}, t, \varnothing) + \alpha \cdot \epsilon_\theta(z_t, t, \mathcal{C_\text{edit}})$ }

   \ENDIF
   \ENDFOR
   \STATE {\bfseries Return} $x_\text{edit} \leftarrow \mathcal{D}(z_0)$

\end{algorithmic}
\end{algorithm}

\section{Experiments}
\subsection{Implementation Details}\label{sec:implementation_details}

In our experiments, we utilized Stable Diffusion V1.5~\cite{rombach2022high} as the foundational model.
We conducted image inversion and generation using DDIM sample steps ($T=50$) per image. 
After extensive empirical evaluation, we set the classifier-free guidance scale to $m=7$.
This value was chosen to optimize the trade-off between maintaining the structural integrity of the original image and ensuring precise semantic alignment with the text prompts, following prior studies~\cite{ho2022classifier}.
Lower guidance scales resulted in insufficient incorporation of text-based modifications, while higher scales led to excessive alterations that compromised image fidelity.
All computations were performed on an NVIDIA RTX 4090 GPU, with null prompt optimization taking an average of approximately 1 minute per image or less.

\subsection{Datasets}

The experiments were conducted using a high-quality dataset of images, selected to represent a diverse range of scenes and objects from the Oxford-IIIT Pets datasets~\cite{parkhi12a}, DreamBooth~\cite{ruiz2023dreambooth}, AFHQ dataset~\cite{choi2020starganv2}, CelebA~\cite{liu2015faceattributes} COCO dataset~\cite{chen2015microsoft} and Custom Diffusion~\cite{kumari2023multi}.

\subsection{Qualitative Results}
\label{qualitative}

We evaluate the effectiveness of our proposed framework across reference-guided and text-guided image editing tasks. Figures~\ref{fig:reference_change}, \ref{fig:afhq}, and \ref{fig:text_edit_result} showcase how our approach maintains structural integrity while achieving high-fidelity attribute transfer.

\subsubsection{Reference-Guided Image Editing}

\begin{figure}[t]
    \centering
    \includegraphics[width=\linewidth]{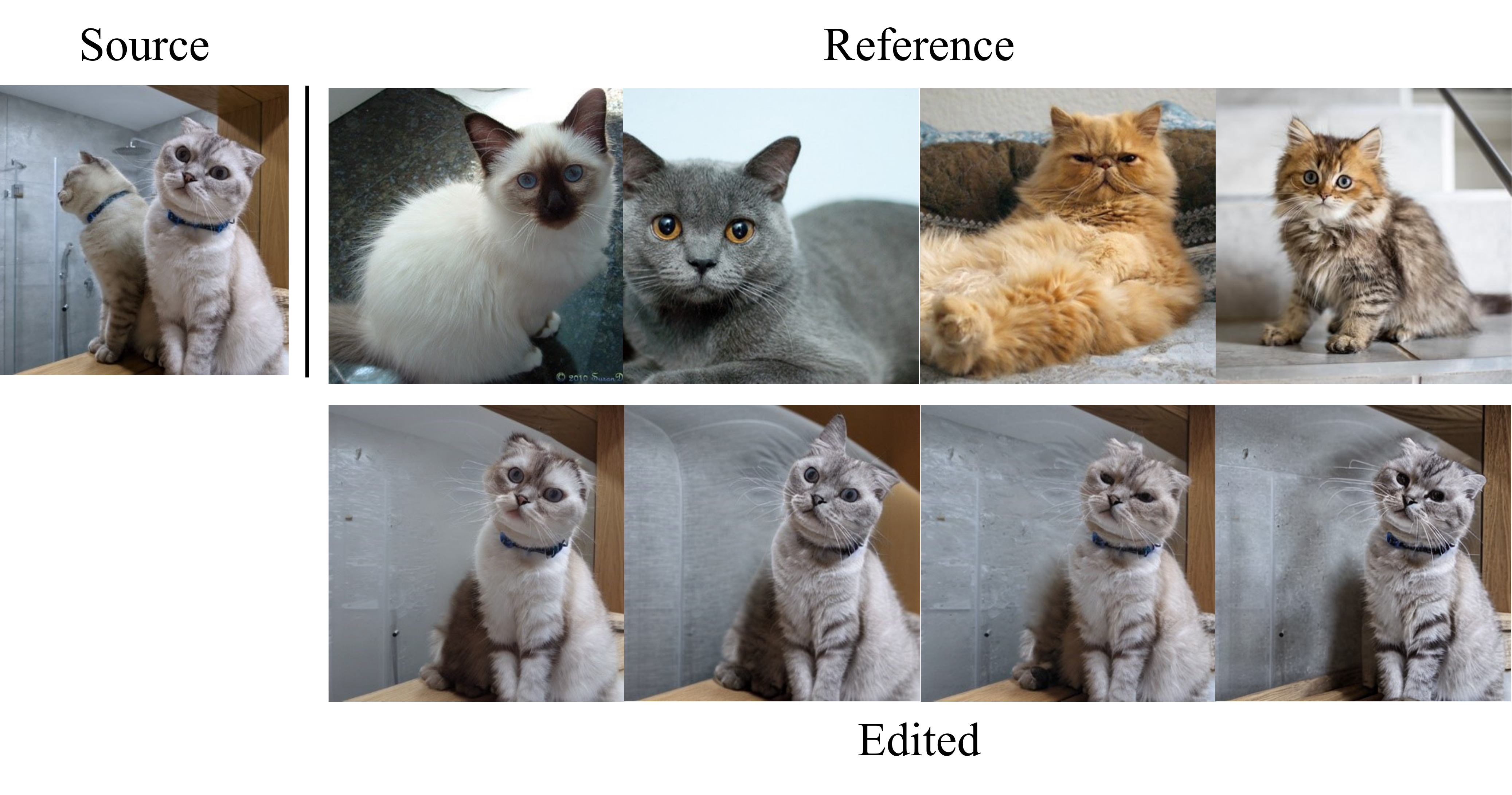}
    \caption{The outputs show that when the source image is fixed, the proposed method accurately integrates stylistic attributes from various reference images while preserving the source’s structure.}
    \label{fig:reference_change}
\end{figure}

Figure~\ref{fig:reference_change} illustrates the effectiveness of our framework in reference-guided image editing, where attributes such as texture, expression, and color are transferred from a reference image to the source image. Unlike conventional methods that often introduce unwanted deformations or structural inconsistencies, our approach ensures that the spatial structure of the source image remains intact. By leveraging self-attention on the reference image and stage-wise attribute injection, our method enables seamless integration of fine-grained attributes.

\begin{figure*}
    \centering
    \includegraphics[width=0.8\linewidth]{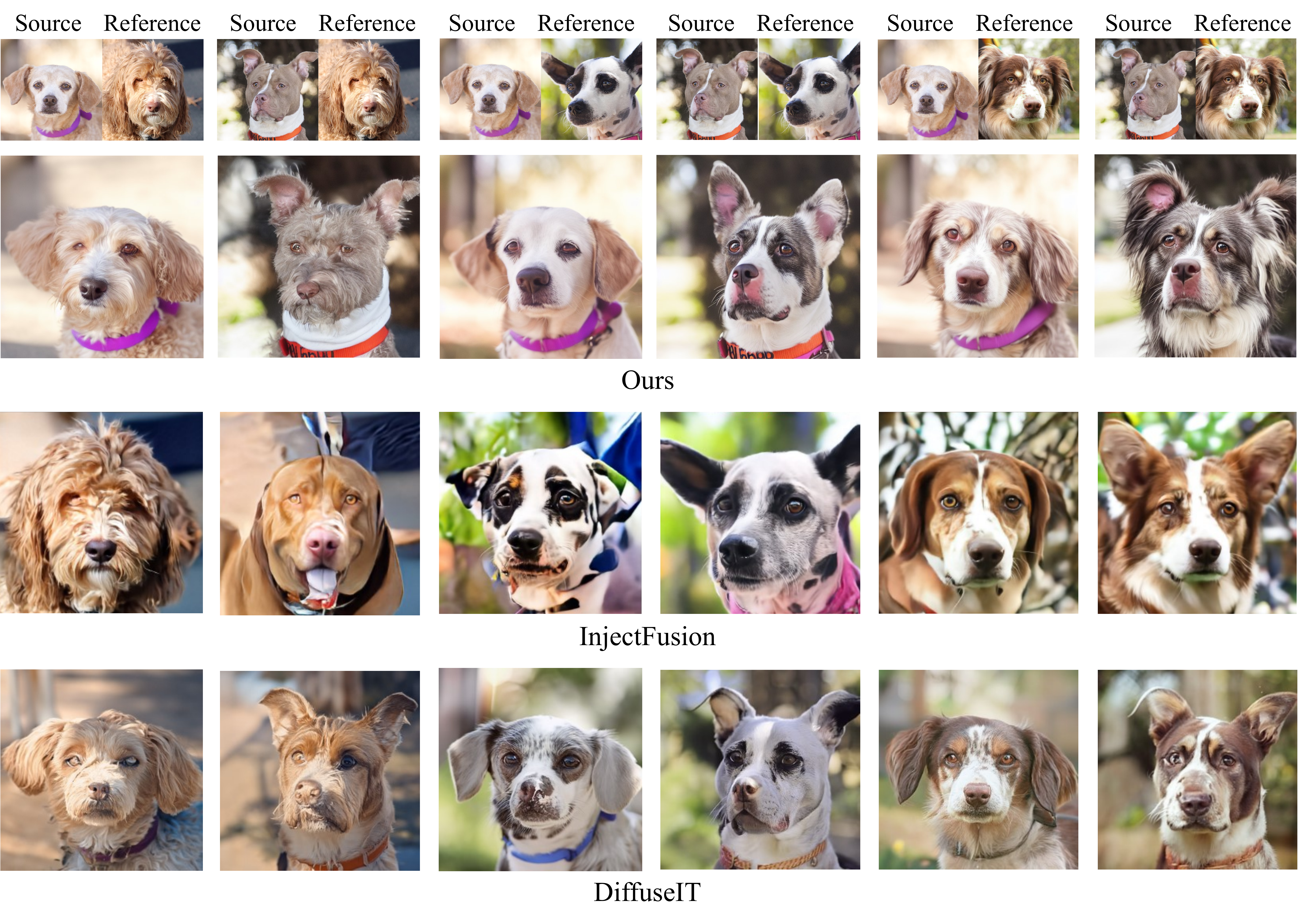}
    \caption{\textbf{Qualitative comparison of generated results across different methods on the AFHQ dataset.} The first and second column contains the source and reference images. The subsequent columns display outputs from (left to right) Ours, InjectFusion, and DiffuseIT.}
    \label{fig:reference_result_afhq}
    \label{fig:afhq}
\end{figure*}

To evaluate the performance of the proposed method, we conducted a comparative analysis of the AFHQ dataset~\cite{choi2020starganv2}, which focuses on diverse dog breeds. Figure~\ref{fig:reference_result_afhq} presents qualitative results that compare our approach with baseline methods, including InjectFusion~\cite{jeong2024training} and DiffuseIT~\cite{kwon2022diffusion}.

Our method preserves the original structure of the input image, including facial features, ear shapes, and the overall pose of the dogs, while other methods tend to introduce artifacts or distortions. 
The proposed approach retains high-quality visual details such as fur texture and sharpness. 
Competing methods, particularly InjectFusion, occasionally produce overly smooth or unrealistic outputs. 
While other methods show inconsistencies in fur color and background blending, our approach ensures consistent color alignment with the reference.
Overall, the proposed model achieves state-of-the-art results in terms of perceptual quality and structural coherence, outperforming existing methods for zero-shot image-to-image translation tasks.

\subsubsection{Comparison with Baseline Methods}

To validate the robustness of our approach, we compare it with InjectFusion~\cite{jeong2024training} and DiffuseIT~\cite{kwon2022diffusion} on the AFHQ dataset in Fig.~\ref{fig:afhq}. While all methods attempt to transfer reference attributes, our approach consistently preserves key structural features, such as facial expressions, ear shape, and pose. InjectFusion tends to produce overly smooth outputs that lack fine-grained texture details, while DiffuseIT often results in structural distortions, particularly in features like fur patterns. In contrast, our model achieves a better balance between structural fidelity and semantic adaptation, making it more reliable for zero-shot image-to-image translation.

\begin{figure}
    \centering
    \includegraphics[width=\linewidth]{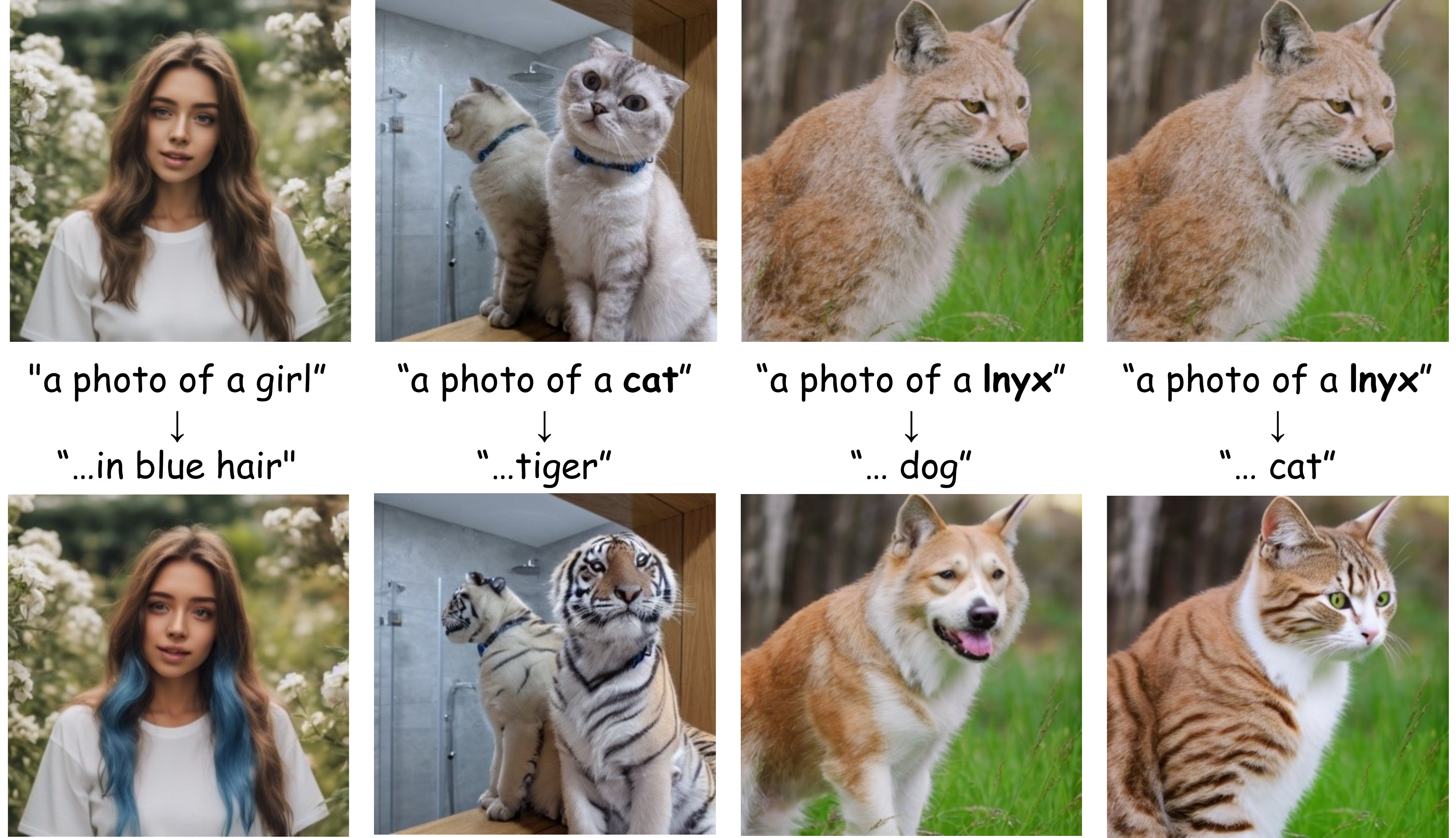}
    \caption{\textbf{Results of text-guided image editing using the proposed method}. The framework seamlessly integrates semantic modifications (e.g., adding a hat, changing hair color, or transforming species) based on text prompts, while preserving the structural and stylistic attributes of the source images.}
    \label{fig:text_edit_result}
\end{figure}

\begin{table*}[h]
\caption{\textbf{Quantitative evaluation of zero-shot image-to-image translation methods on the AFHQ dataset using Palette and CLIP metrics.} Palette (src/ref) measures the perceptual discrepancy between the source/reference image and the generated image, with lower values indicating better performance. CLIP (src/ref) evaluates semantic consistency, with higher values representing better alignment.}
\label{tab:quantitative}
\vskip 0.15in
\begin{center}
\begin{small}
\begin{sc}
\begin{tabular}{lcccc}
\toprule
Metric & Palette(src) & Palette(ref) & CLIP(src) & CLIP(ref) \\
\midrule
Ours & 0.4846 & 1.0237 & \textbf{0.8883} & \textbf{0.9141}  \\
DiffuseIT & 0.3425 & 0.7242 & 0.8559 & 0.9045  \\
InjectFusion & 0.5703 & 0.7782 & 0.7965 & 0.8863  \\
\bottomrule
\end{tabular}
\end{sc}
\end{small}
\end{center}
\vskip -0.1in
\end{table*}

\subsubsection{Text-Guided Image Editing}

Figure~\ref{fig:text_edit_result} showcases the performance of the proposed method for text-guided image editing. The top row presents the source images, and the bottom row illustrates the edited outputs based on specific text prompts. The proposed method effectively applies the semantic modifications specified in the text while preserving the structural integrity and stylistic coherence of the source images.

The first example modifies hair color to "blue," accurately reflecting the prompt while preserving the original face structure and expression.
In the second example, the source image of a cat is transformed into a "tiger," maintaining the original pose and spatial arrangement while adopting the texture and patterns of a tiger.
The third and final example transforms a lynx into other animal forms like a dog or a domestic cat, showcasing the method's ability to integrate diverse semantic changes while maintaining structural consistency.

These results demonstrate the flexibility and robustness of the proposed framework for text-guided edits across various scenarios, achieving high fidelity in both semantic adaptation and structural preservation.

\subsection{Quantitative Results}

As shown in Table~\ref{tab:quantitative}, the results present a quantitative comparison of zero-shot image-to-image translation methods on a subset of 12 images generated from the AFHQ dataset, focusing on diverse dog breeds. The results are evaluated using Palette and CLIP metrics.
We provide a quantitative comparison of our method with DiffuseIT~\cite{kwon2022diffusion} and InjectFusion~\cite{jeong2024training} on two types of metrics.

Our method achieves a Palette (ref) score of 1.0237, which is slightly higher than DiffuseIT (0.7242) and InjectFusion (0.7782). While our method shows a marginally higher discrepancy, as seen in the CLIP scores, it prioritizes semantic consistency. 
In terms of Palette (src), our approach achieves a moderate score of 0.4846, balancing the source image structure while effectively transferring attributes from the reference. DiffuseIT achieves the best score (0.3425), indicating that it retains the source structure most effectively.

Our method achieves the highest CLIP (src) score of 0.8883, highlighting superior semantic alignment with the source image.
Similarly, for CLIP (ref), our method scores 0.9141, surpassing both InjectFusion (0.8863) and DiffuseIT (0.9045), demonstrating effective semantic attribute transfer from the reference image.

The proposed method demonstrates state-of-the-art performance across diverse images from the AFHQ dataset, achieving an optimal trade-off between perceptual fidelity (Palette Metric) and semantic consistency (CLIP Metric). These results emphasize the robustness and generalizability of our approach in zero-shot image-to-image translation scenarios.

\section{Ablation Study}

\subsection{Effect of Null-Text Inversion and Injection Mechanisms}

\begin{figure}[h]
    \centering
    \includegraphics[width=\linewidth]{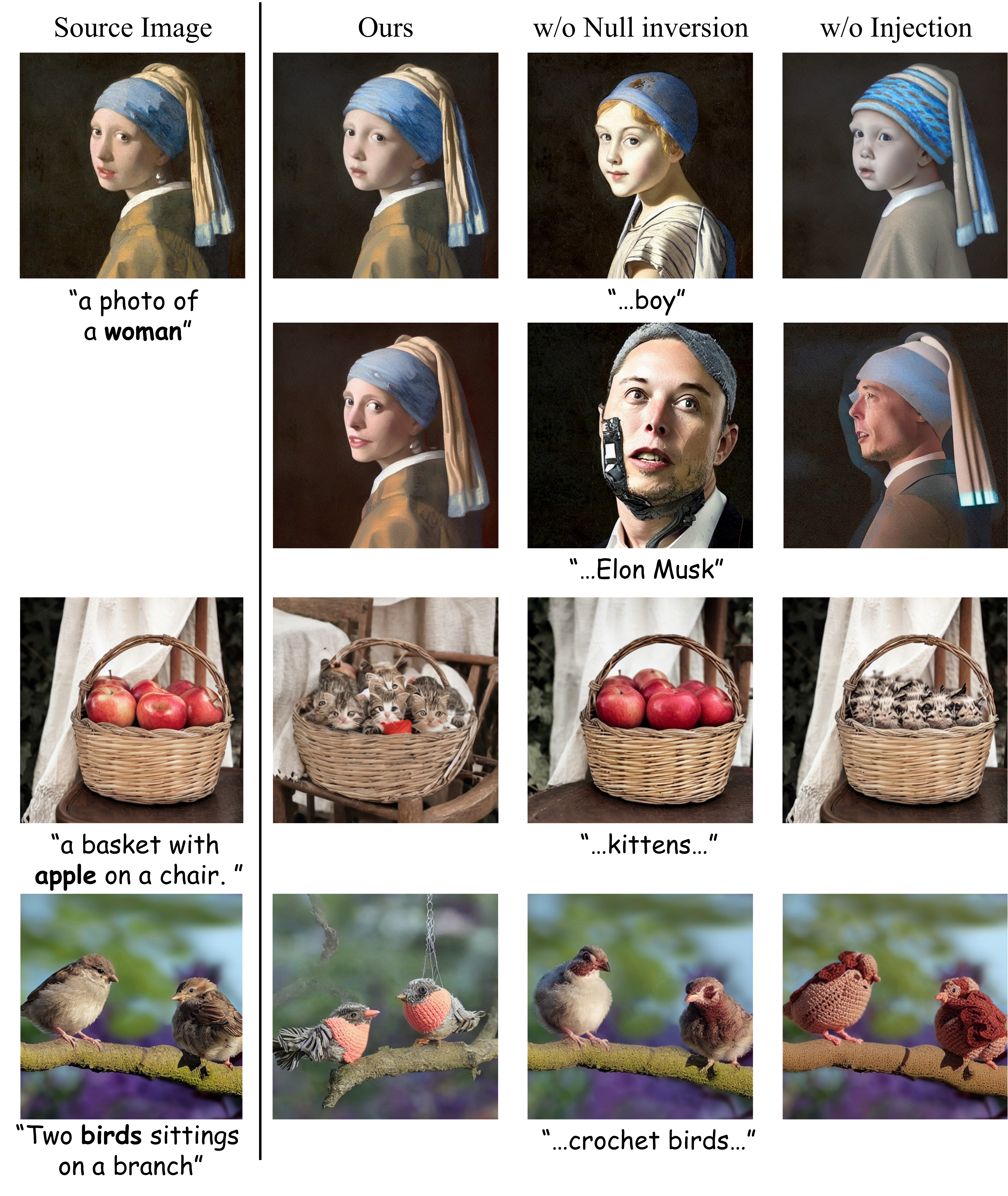}
    \caption{\textbf{Ablation study on the effects of Null-Text Inversion and stage-wise injection mechanisms.} When Null-Text Inversion is removed (w/o Null Inversion), the model fails to maintain structural integrity, leading to distorted outputs. When the stage-wise injection mechanisms are removed (w/o Injection), attribute transfer is incomplete or inconsistent. Our full model (Ours) successfully preserves structure while integrating fine-grained modifications, demonstrating the necessity of both components for high-quality zero-shot image editing.}
    \label{fig:ablation}
\end{figure}

To analyze the contribution of key components in our framework, we conduct an ablation study by removing Null-Text Inversion and stage-wise injection mechanisms separately. The results of this analysis are presented in Fig.~\ref{fig:ablation}.

When Null-Text Inversion is removed, the model struggles to maintain the structural integrity of the source image. As shown in Fig.~\ref{fig:ablation}, the transformation from “a woman” to “Elon Musk” results in unnatural distortions, and object-level modifications such as “birds to crochet birds” exhibit artifacts and inconsistencies. These results confirm that Null-Text Inversion is essential for anchoring spatial structure, preventing unnecessary deformations while allowing controlled modifications.

When stage-wise injection mechanisms are removed, the reference-based modifications become inconsistent. The edited images fail to integrate local attributes properly, leading to unnatural blending or incomplete transformations (e.g., "basket with apple on a chair" retains the original chair texture instead of blending naturally). This highlights the importance of our two-stage injection strategy, which first anchors structural features and later infuses fine-grained details.

Our full model successfully combines structural preservation and fine-grained attribute control, producing high-quality transformations without unwanted distortions. This study validates that both Null-Text Inversion and stage-wise injection mechanisms are essential for robust and coherent zero-shot image editing.

\subsection{Effect of Attribute Injection Timing}

When attribute injection occurs too early in the denoising process, the model tends to overfit to the new textual prompt, resulting in substantial structural deformations. For instance, in the "clock in the Grand Canyon" example, injecting attributes at early timesteps causes severe warping that distorts the original geometric layout. Conversely, if the injection is delayed excessively, the semantic transformation becomes too weak. This is exemplified in the "cat to tiger" case, where the image fails to exhibit distinctive tiger-like textures, resulting in inconsistent appearance.

Our proposed method introduces attribute injection at later timesteps to strike a balance between semantic adaptation and structural preservation. By applying attribute injection after the structural layout has stabilized, we enable fine-grained modifications that preserve the core spatial features of the source image. The results confirm that careful control of injection timing significantly improves both perceptual quality and semantic consistency in zero-shot editing.

\begin{figure}[h]
    \centering
    \includegraphics[width=\linewidth]{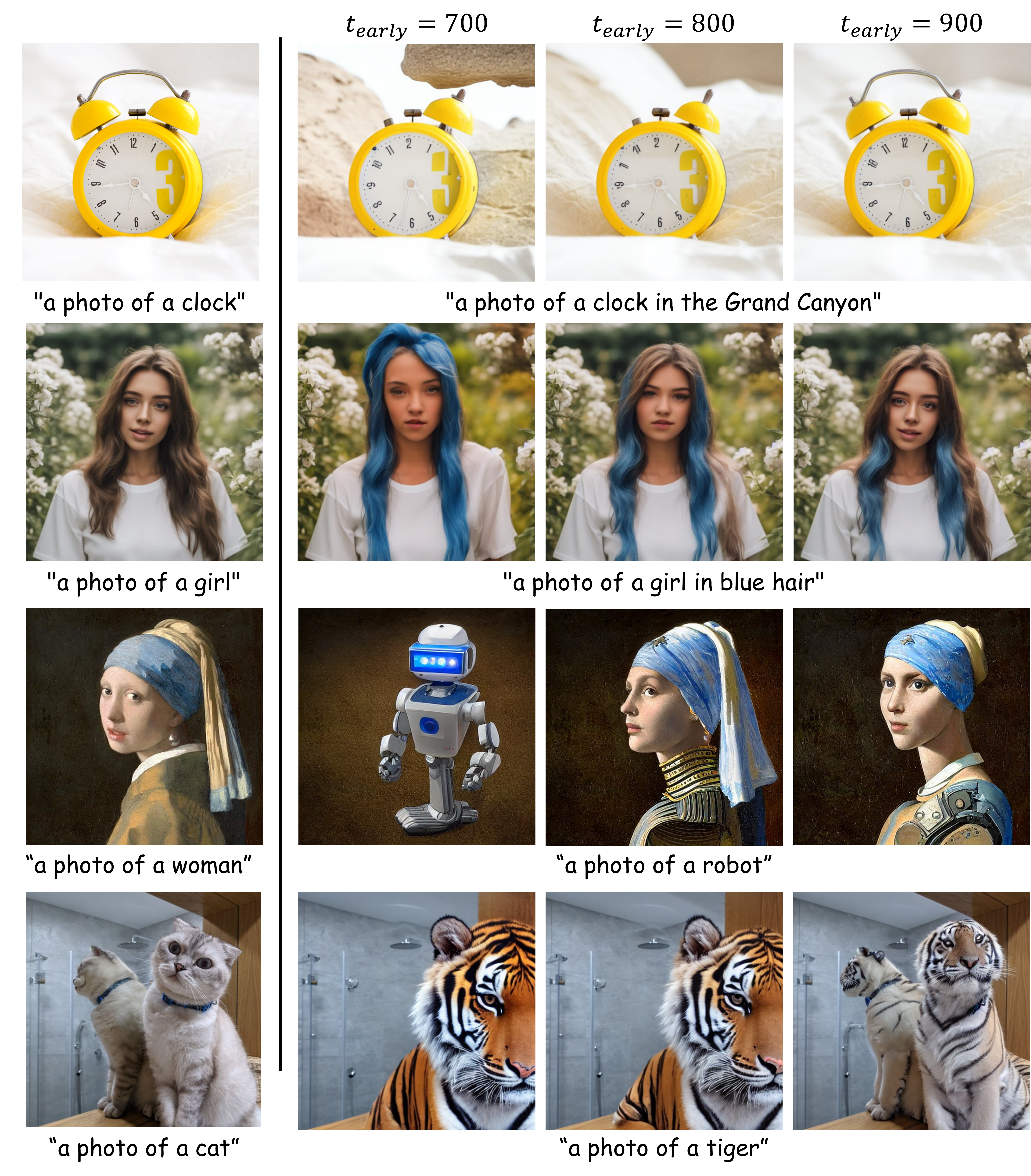}
    \caption{\textbf{Effect of attribute injection timing in text-guided image editing.} When injection occurs too early, the structural integrity of the source image is lost, leading to distortions. When applied too late, the semantic transformation is incomplete. Our approach optimally injects attributes in later timesteps, achieving both structural preservation and faithful text-guided modifications.}
    \label{fig:adain}
\end{figure}

\section{Conclusion}
We introduced a zero-shot image editing framework that unifies text-guided and reference-guided editing within a diffusion-based architecture, leveraging optimized null-text embeddings and a stage-wise latent injection strategy to achieve structural preservation and fine-grained attribute transfer without fine-tuning. Self-attention on the reference image maintains shape consistency in early timesteps, while attribute injection in later timesteps enables seamless integration of semantic details. Extensive experiments across various editing tasks, including expression transfer, texture transformation, and style infusion, demonstrate that our method outperforms InjectFusion and DiffuseIT in perceptual quality, semantic consistency, and structural coherence, with ablation studies validating the necessity of null-text inversion and staged injections. While our approach remains sensitive to extreme domain shifts, future work will explore multi-modal conditioning, such as incorporating sketch or depth guidance, to further enhance adaptability and control. Our framework establishes a new benchmark in zero-shot image editing, demonstrating the potential of diffusion models for structure-preserving, high-fidelity transformations.

\section{Supplementary Materials}
\subsection{Additional Image Sources}
\begin{itemize}
    \item Birds on a branch: \url{https://pixabay.com/photos/sparrows-birds-perched-sperlings-3434123/}
    \item Basket with apples: \url{https://unsplash.com/photos/4Bj27zMqNSE}
    \item Cat sitting next to a mirror: \url{https://null-textinversion.github.io/}
    \item A photo of a girl: \url{https://ip-adapter.github.io}
\end{itemize}

{
    \small
    \bibliographystyle{ieeenat_fullname}

}


\end{document}